\documentclass[preprint,11pt,review]{elsarticle}

\usepackage{amssymb,amsmath,paralist}
\usepackage{xfrac}
\usepackage{lipsum}
\usepackage[english]{babel}
\usepackage[utf8]{inputenc}
\usepackage{graphicx,xcolor,adjustbox}
\usepackage[font=small,labelfont=bf,tableposition=top]{caption}
\usepackage[font=footnotesize]{subcaption}
\usepackage{amsfonts}
\usepackage{mathtools}
\usepackage{etoolbox}
\usepackage{mdframed}
\usepackage{enumitem}

\usepackage{algorithm}
\usepackage{listings}
\usepackage{algpseudocode}

\usepackage{bbm}
\usepackage{multirow}
\usepackage{rotating}
\usepackage{float}
\usepackage{subfiles}

\usepackage{lineno}

\usepackage{nicematrix}
\usepackage[all,cmtip]{xy}
\usepackage{tikz}
\usetikzlibrary{calc,fit}
\usepackage{float}
\usepackage{scrextend}
\usepackage[]{upgreek}
\usepackage{tabularx}
\usepackage{booktabs}

\newcommand{\dsp}{\displaystyle}
\newcommand{\bfs}{\bfseries}

\newcommand{\re}{\mathbb{R}}
\newcommand{\nn}{\mathbb{N}}

\newcommand{\lpar}{\left ( }
\newcommand{\rpar}{ \right )}
\newcommand{\lset}{\left\lbrace}
\newcommand{\rset}{\right\rbrace }
\newcommand{\lbr}{\left [  }
\newcommand{\rbr}{  \right ] }

\newcommand{\tend}[1]{\, \underset{#1 \to \infty}{\longrightarrow}  \, }

\newcommand{\algName}{Doubly Stochastic Mean Shift}
\newcommand{\msop}{\mathcal{S}}  
\newcommand{\shv}{\mathbf{m}_{h}} 
\newcommand{\kde}{D} 
\newcommand{\shortname}{SMS\,}
\newcommand{\dsm}{DSMS\,}
\newcommand{\mshift}{MS\,\,}
\newcommand{\truedensity}{f_{X}}
\newcommand{\kernprof}{\kappa}
\newcommand{\bms}{BMS\,}
\newcommand{\point}{\mathbf{x}}
\newcommand{\state}{\mathcal{X}}

\newcommand{\ridx}[1]{i_{#1}}
\newcommand{\neighb}[1]{N_{h} (#1)}

\newcommand{\SmallSubg}[1]{\mathcal{A}_{#1}}

\newtheorem{theorem*}{{\bfs \text{Theorem}}}
\newtheorem{def*}{{\bfs \text{Definition}}}
\newtheorem{lem*}{$\textbf{Lemma}$}
\newtheorem{pf*}{$\textit{Proof}$}
\newtheorem{prop*}{\textbf{Proposition}}
\newtheorem{cor*}{{\textbf{Corollary}}}
\newtheorem{rem*}{{\bfs \text{Remark}}}
\journal{Signal Processing}

\newcommand{\dobib}{ 
    \bibliographystyle{model5-names}\biboptions{authoryear}
    \bibliography{mybib} 
}

\usepackage{newfloat}
\DeclareFloatingEnvironment[
    fileext=los,
    listname={List of Algorithms},
    name=Algorithms,
    placement=tbhp,
    within=none
    ]{algo}
    
\newcommand{\prooflocation}{appendix}

\newif\ifshortVersion
\shortVersionfalse

\newif\iflongVersion
\longVersiontrue

\graphicspath{{./figures/}{./}}

\begin{document}

\renewcommand{\dobib}{}

\begin{frontmatter}



\title{Doubly Stochastic Mean-Shift Clustering}


\author[1]{Tom Trigano}
\ead{thomast@sce.ac.il}
\address[1]{Shamoon College of Engineering, Department of Electrical Engineering, Ashdod, Israel}

\author[2]{Yann Sepulcre}
\ead{yanns@mail.sapir.ac.il}
\address[2]{Sapir Academic College, Department of Engineering, Sderot, Israel}

\author[3,4]{Itshak Lapidot}
\ead{itshakl@afeka.ac.il}
\address[3]{Afeka Tel-Aviv Academic College of Engineering, School of Electrical Engineering, Israel}
\address[4]{Avignon University, LIA, France}

\begin{abstract}

Standard Mean-Shift algorithms are notoriously sensitive to the bandwidth hyperparameter, particularly in data-scarce regimes where fixed-scale density estimation leads to fragmentation and spurious modes. In this paper, we propose Doubly Stochastic Mean-Shift (DSMS), a novel extension that introduces randomness not only in the trajectory updates but also in the kernel bandwidth itself. By drawing both the data samples and the radius from a continuous uniform distribution at each iteration, DSMS effectively performs a better exploration of the density landscape. We show that this randomized bandwidth policy acts as an implicit regularization mechanism, and provide convergence theoretical results. Comparative experiments on synthetic Gaussian mixtures reveal that DSMS significantly outperforms standard and stochastic Mean-Shift baselines, exhibiting remarkable stability and preventing over-segmentation in sparse clustering scenarios without other performance degradation.
\end{abstract}


\end{frontmatter}


\section{Introduction}
\label{sec:Introduction}
Numerous applications of signal processing (e.g. speaker diarization, image segmentation) involve a clustering step. In order to perform it, clustering algorithms have been proposed and investigated, among which the $k-$means~\citep{lloyd_least_1982}, spectral clustering~\citep{shi_normalized_2000,ng_spectral_2002}, DB-SCAN~\citep{ester_density_1996}, the well-known \textit{Mean-shift} (\mshift)~\citep{Fukunaga1975} and its variant the \textit{blurring mean-shift} (\bms)~\citep{Cheng_1995}. Both \mshift and \bms algorithms can be considered as ``deterministic'' iterative procedures aiming to find local maximizers of an objective function, since they do not involve any random selection of points to perform their update rule. 

In a recent publication~\citep{lapidot_stochastic_2025}, the authors introduced the Stochastic Mean-Shift (\shortname), where the data sample points are selected randomly and moved by one iteration. It can still be considered to be a blurring process since the updated points are never replaced to their original locations and keep moving from their current position; at each step any sample point can be picked up for update. The algorithm proved to be convergent almost surely, as the gradient steps asymptotically tend to zero. This algorithm was successfully applied for speaker clustering tasks, where it outperformed \mshift in many cases. Though \shortname represents an improvement of the ordinary \mshift, its properties remain to be investigated in depth. Indeed, stochastic algorithms can prove to be more efficient than deterministic methods for optimization procedures \citep{delmoral_modeles_2014}. 

One of the shortcomings of \shortname lies in the fact that the kernel used has a fixed compact support. A fixed kernel size in most of mean-shift procedures introduces substantial limitations, as it is related to a data homogeneity assumption that does not hold in many setups such as high-dimensional cases. In dense regions, a too large kernel oversmooths the underlying structure, blurring fine-scale modes and potentially merging distinct clusters that should remain separated. Conversely, in sparse regions, a small kernel yields a noisy estimate of the gradient, creating spurious modes and amplifying sampling variability. Fixed bandwidth kernels with radial symmetry also fail to account for anisotropic structures, where the relevant scale differs across directions and locations, thus preventing the algorithm from adapting to elongated or curved manifolds that frequently arise in practice~\cite{zhang_kernel_2021}. It is therefore of great theoretical and practical interest to further investigate other strategies, which could significantly improve the stability and fidelity of mean-shift–based clustering and mode-seeking~\cite{duong_nearest_2016}.

This paper attempts to modify the \shortname algorithm to also select a random bandwidth to be used for the position update of the randomly selected point. Besides preventing partly convergence to false cluster centers, the hypothesis formulated is that random modifications of the kernel bandwidth in \shortname allows data outliers to converge to actual modes, thus reducing the final number of obtained clusters. Randomization of fixed parameters, such as learning rates, has shown to be beneficial to several learning algorithms, e.g.~\cite{shim_adaptive_2023} and~\cite{mamalis_stochastic_2023}. 

We present the improvement of \shortname, called \algName (\dsm), in section~\ref{sec:Doubly_SMS}, theoretical results are presented and discussed in~\ref{sec:theory_with_proofs}, and present results obtained on synthetic datasets in section~\ref{sec:Experiments}. For a better comparison with \shortname, we used the same synthetic datasets as in~\cite{lapidot_stochastic_2025}.  
%
Proofs of the presented results are detailed in the \prooflocation.

\section{Doubly Stochastic mean-shift}
\label{sec:Doubly_SMS}

The following section presents a stochastic version of \mshift based on
random width kernel. 
Table~\ref{tab:notations} summarizes the notations used in this paper.

  \begin{table*}[ht]
    \centering
    \caption{Summary of the main notations used throughout the paper}
    \maxsizebox{0.98\linewidth}{!}{%
      \begin{tabular}{ccp{0.95\linewidth}}
    \toprule
    Notation & \phantom{abcde} & Significance \\
    \midrule
    \phantom{aaaaaaa}         & General notations & \\
    $[n]$ & & the set of integers $\{1 , \ldots , n \} $ \\
    $\re_{+}$ & & the set of non negative reals $[ 0,\infty )$ \\
    $\| \mathbf{u} \| \, , \ \mathbf{u} \in \re^{n\times 1} $ & & the $\ell_2 $ norm
    of $\mathbf{u}$ \\
    \midrule
    \phantom{aaaaaaa}             & Specific notations & \\
    $x_1,\ldots x_J$ & & The $J$ data points \\
    $K(\mathbf{u}) = k\lpar  \| \mathbf{u} \|^2 \rpar \ ; \ G(\mathbf{u}) = - k^{\prime}\lpar\| \mathbf{u}\|^2 \rpar $ & & density kernel $K$ on $\re^n $ associated to the profile $k:[0,\infty) \to \re_{+}$ \, ; \, "weight" function on $\re^n $ \\
    $\neighb{x}, \, \, x \in \re^d$ & & subset of $[n]$ of the neighboring indices $\left\{ {i \, :\left\| {x  - {\point_i}} \right\| < h} \right\}$ \\
    $\state = [\point_{i}]_{i\in [n]} \in \re^{n d} $ & & a state, i.e. a column vector made of the ordered collection of $n$ points each one in $\re^d $ \\
    $L_h (\state)$ & & the cost function $\sum\limits_{1\leq i\leq j \leq n} K \lpar \frac{\point_i - \point_j }{h}\rpar $ at state $\state$, depending on the bandwidth $h$ \\
    $\ridx{k}$ & & the index $i \in [n]$ randomly chosen at the $k^{th}$ step
    of \dsm \\
    $h_k $ & & the BW at the $k^{th}$ step of \dsm \\
    $\nabla_{i} L_h $ & & partial gradient $\nabla_{\point_i } L_h $ of $L_h $ w.r.t the variable $\point_i $ \\
    $\nabla^{(k)}_{h} \ ; \ \nabla_{i\,;\,h}^{(k)} $ & &  gradient of $L_h$ at step $k$ of \dsm \, ; \,  partial gradient $\nabla_{i} L_h $ at step
    $k$ of \dsm \\
    $\| \mathbf{U} \| \, , \ \mathbf{U} = [\mathbf{u}_{1}^{T} , \ldots , 
    \mathbf{u}_{n}^{T} ]^{T} \, , \, \, \mathbf{u}_i \in \re^{d\times 1} $ & & 
    $ \max\limits_{i \in [n]} \| \mathbf{u}_i \|_2 $: maximal component wise $\ell_2 $ norm value \\
   $[h_{min},h_{max}]$ & & the interval of considered bandwidths \\
     \bottomrule
  \end{tabular}}
  \label{tab:notations}
\end{table*}


First, we provide details on the specific kernels to be considered in this work. These assumptions will be further used in the theoretical proofs.

\begin{def*}
    \begin{enumerate}
        \item We call a profile function a mapping $k: [0,1] \to \re_{+} $ that is $C^1 $, non-increasing, convex and such that $k^{\prime}(t)<0 $ for all $t \in [0,1)$.
  \item a multivariate function which can be written as $K( \mathbf{u} ) = k\lpar  \| \mathbf{u} \|^2 \rpar$ for all $\mathbf{u} \in \re^d$ 
  will be referred to as a kernel with profile function $k$.
    \end{enumerate}
\end{def*}
Typical examples of such profiles and associated kernels include $k(t)= \lpar 1-t \rpar_{+}^{\alpha}$ for $\alpha=2,3,4$, which lead respectively to the Bi-, Tri-, and Quadweights kernels
 $ K( \mathbf{u} ) = \lpar 1- \| \mathbf{u} \|^2  \rpar_{+}^{\alpha} $, respectively. 

For any sample set $\{ \point_i \}_{i \in [n]}$ with $\point_i \in \re^{d}, i=1\ldots n $, stemming from some unknown multi-modal density $\truedensity$ defined on $\re^d $, given a kernel $K$ and a positive bandwidth parameter $h>0$, the kernel density estimate (KDE) of $\truedensity$ associated with $\{ \point_i \}_{i \in [n]}$, denoted by $\kde$, verifies
 \begin{align}\label{eq:KDE_def}
  \forall \, x \in \re^d \ \,    \kde (x) & \propto 
  \frac{1}{n h^d }\sum\limits_{i \in[n]} \, K\left ( \frac{1}{h}\, (x - \point_{i} )\right ).
 \end{align}
The {\it{mean-shift}} algorithm~\citep{Fukunaga1975,Senoussaoui2014,Comaniciu2002} is a well known clustering method, based on finding the 
local maxima of \eqref{eq:KDE_def}.
 Since all the considered kernels have compact support by definition of the profile functions, this property is shared by any KDE defined in \eqref{eq:KDE_def}. For the rest of this paper, 
 we shall refer to the element $\state=[\point_{1}^\top, \ldots , \point_{n}^{\top} ]^{\top} \in \re^{d \, n}$ as the \textit{state} defined by this sample, and interpret any iterative clustering algorithm as a chain of transformations of $\re^{d\,n}$ applied to the initial state $\state^{(0)}$ given by the raw sample until we reach convergence (in a sense to be precised). A state shall be introduced further shortly as 
 $\state = [\point_{i}]_{i \in [n]}$ for the sake of simplicity, without emphasizing further the implicit vector structure.
 
\subsection{Blurring mean-shift and Stochastic mean-shift}
\label{subsec:stochasticMS}
For any state $\state$, we shall consider the {\it{mean-shift operator}} $\msop_{h} : \mathbb{R}^d \to \mathbb{R}^d $ which maps a point $x \in \mathbb{R}^d $ to its new position as
 \begin{equation}\label{eq:meanshiftOperator}
 \msop_{h} (x ; \state) = \frac{\displaystyle \sum\limits_{i \in [n]} G\left ( \frac{x - \point_i }{h}\right )\, \point_i  }{\displaystyle \sum\limits_{i \in [n]} G\left ( \frac{x - \point_i }{h}\right )} \ \ \, , \ \ \, 
 \end{equation}
where $G(\mathbf{u}) = - k^{\prime}\lpar\| \mathbf{u}\|^2 \rpar$ defines the non-negative {\it{weight function}} associated with $K$.
Note that the denominator in \eqref{eq:meanshiftOperator} is always positive since $-k^{\prime}(0)>0$. Note also that \eqref{eq:meanshiftOperator} is effectively computed on the {\textit{$\state-$neighborhood}} of $x$ which is defined as $$\neighb{x} = \lset i \in [n] \, | \, G\lpar \frac{x - \point_{i} }{h}\rpar \neq 0 \rset =  \lset i \in [n] \, | \, \| x - \point_{i} \| < h \rset, $$  Since $K$ has a radial symmetry, 
 the \textit{shift vector} of $x$ is defined as
 \begin{equation}\label{eq:shift_definition}
     \shv (x ; \state) = \msop_{h} (x ; \state) - x \ \ ;
 \end{equation}
 it can be easily verified that $ \shv (x ) $ is indeed proportional to the gradient of 
 $\kde$ taken at $x$. 

The Blurring mean-shift (\bms) is a kernel-based iterative method for data clustering. 
  Starting from the initial state $\state^{(0)} \in \re^{d n}$ defined by the original data sample, for any $k \in \mathbb{N}_0 $ the $k^{th}$ step of the \bms consists in updating the current state $\state^{(k)}$ to the new state $\state^{(k+1)}$ by applying the mean-shift operator defined as follows:
 \begin{align}\label{eq:kth_step_bms}
     \forall \, i \in [n] \ \ \ & \point_{i}^{(k+1)} = \frac{\sum\limits_{j=1}^n G\lpar \displaystyle \frac{\point^{(k)}_i - \point^{(k)}_j  }{h }\rpar \, \point^{(k)}_j }{\sum\limits_{j=1}^n G\lpar \displaystyle \frac{\point^{(k)}_i - \point^{(k)}_j  }{h} \rpar }
 \end{align}
 In other words, we apply to each sample point the mean-shift operator to find its new location, but by plugging in \eqref{eq:meanshiftOperator} the current state $\state^{(k)}$.


 From a mathematical point of view, the BMS algorithm can be seen as an iterative procedure aimed at converging to a local maximizer of the following non-negative objective function
\begin{align}\label{def:L_function}
    \forall \, \state=[\point_{1}^{\top} , \ldots , \point_{n}^{\top} ]^{\top} \in \re^{d\,n} \ \ \, L(\state) & = \sum\limits_{1 \leq i \leq j \leq n } K\lpar \frac{1}{h} (\point_i - \point_j )\rpar  \ \ \ ;
\end{align}
Indeed one can reformulate \eqref{eq:kth_step_bms} as a "weighted" gradient step as follows: 
 \begin{align}\label{eq:step_bms_with_gradient}
      \state^{(k+1)} = \state^{(k)}  + \frac{h^{2}}{2} \sum\limits_{i=1}^n \frac{1}{\sum\limits_{j=1}^n G\lpar \displaystyle\frac{\point_{i}^{(k)} - \point_{j}^{(k)}}{h}\rpar} \nabla_{\point_{i}} L (\state^{(k)})
 \end{align}
 where $\nabla_{\point_{i}} L (\state)$ denotes the  gradient component of $L$ related to the $\point_{i}$ direction.
 Any state $\state \in \re^{d n} $ satisfies $\nabla L (\state) = \mathbf{0}_{d n} $ if and only if it is a fixed point of the transformation \eqref{eq:kth_step_bms}, i.e.
\begin{align}\label{eq:critical_point}
    \forall \, i\in [n]  \, \ \ \point_i = \frac{\sum\limits_{j=1}^n G\lpar \displaystyle \frac{\point_i - \point_j  }{h }\rpar \, \point_j }{\sum\limits_{j=1}^n G\lpar \displaystyle \frac{\point_i - \point_j  }{h} \rpar }
\end{align}
  The relation between the stationary points of $L$ and our clustering objective is made explicit in a theorem from~\citep{Yamasaki_2024}, which can be written as follows:
  \begin{theorem*}\label{thm:critical_point_charact}
      Assume the profile function $k$ of the kernel $K$ satisfies the assumptions of Definition 1. A state $\state =[\point_{1}^{\top} , \ldots , \point_{n}^{\top} ]^{\top} \in \re^{d \, n} $ satisfies $\nabla L (\state)= \mathbf{0}_{d n}$ iff the following is satisfied:     \begin{align}\label{eq:criticalpoint_geometric_char_UPDATED}
          \forall \, (i,j) \in [n]^2 \ \, j \in \neighb{\point_{i}} \, \iff \, \point_j = \point_i 
      \end{align}
  \end{theorem*}
  In other words, for the specific truncated kernels considered in this study, a critical point of $L$ is a state satisfying an ideal clustering condition: two points either coincide, or are distant from $h$ at least. In the previous theorem, \eqref{eq:criticalpoint_geometric_char_UPDATED} will be later used to prove theoretical results on the effectiveness of the doubly stochastic mean-shift. 
Convergence of the \bms sequence has been thoroughly investigated and proved~\citep{Chen_2015}; theorems establishing convergence rates for different kernel types can be found in~\citep{Yamasaki_2024}. 


The \textit{stochastic mean-shift} algorithm
(\shortname) was presented in~\citep{lapidot_stochastic_2025}. Roughly
speaking, \shortname is a blurring method like \bms but adds some randomness in the shifting process. Its principle is
as follows: starting from an original state ${{\cal X}}^{(0)} = \lbr
\point_{i}^{(0)} \rbr_{i = 1}^{n}$, in each subsequent step an index $i
\in [n]$ is picked up randomly (according to the uniform distribution
and independently of all previous choices), and  the chosen point alone is moved by using the shift operator $\msop_{h}$, by plugging in \eqref{eq:meanshiftOperator} the sample $\lbr
\point_{i}^{(k)} \rbr_{i = 1}^{n}$ considered in its \textit{current state} at step $k$. 
A summarized version of \shortname is described in Algorithm \ref{alg:stochasticMeanShiftClustering}. Convergence diagnosis is
    either based on a maximal number of iterations or when
    $\|\mathbf{x}_i^{(k+1)}-\mathbf{x}_i^{(k)}\|$ is smaller than a
    user-defined threshold. 

As noted in~\cite{lapidot_stochastic_2025} \shortname can be interpreted as some "partial" gradient algorithm: starting from some initial state $\state^{(0)} \in \re^{nd}$, the $k^{th}$ step of \shortname consists in picking randomly an index $\ridx{k} \in \{1,\ldots,n\}$ and defining the $k+1$ state
  as 
  \begin{align}\label{eq:step_sbms_intro}
      \state^{(k+1)} = \state^{(k)} + \frac{h^{2} /2}{\sum\limits_{j=1}^n G\lpar \frac{\point_{\ridx{k}}^{(k)} - \point_{j}^{(k)}}{h}\rpar} \nabla_{\point_{\ridx{k}}} L (\state^{(k)})
  \end{align}
  where $\nabla_{\point_{i}} L (\state)$ denotes the partial gradient of $L$ in the $\point_{i}$ direction. 
Algorithm \ref{alg:stochasticMeanShiftClustering} shows that \shortname has a linear complexity $O(k)$ in the the overall number of draws $k$, and computational advantages related to this property were discussed in~\cite{lapidot_stochastic_2025}. 

\subsection{Doubly stochastic mean-shift}
\label{subsec:Doubly_SMS}
 In the \shortname algorithm considered previously, the \textit{bandwidth} (BW) $h$ to be plugged in the kernel based update formula stays constant. In this work we shall introduce the {\it{Doubly Stochastic mean-shift}} (\dsm) which proceeds as the \shortname , but also picks up randomly a bandwidth $h_{k+1}$ at each step $k$ 
 (according to some specified distribution). The BW $h_{k+1} $ at step $k$ will be chosen inside a fixed positive interval $[h_{min} \, ; \, h_{max}]$ for the following reason: too high values of $h$ would likely expose the chosen point to the influence of too far points belonging to other clusters; and a too low value of $h$ may lead to a poor update (if any). One version of the \dsm is presented in Algorithm \ref{alg:doubly_stochasticMSC} below and shall be investigated in the theoretical section \ref{sec:theory_with_proofs}. This version is indeed one among many others possible, since several ways to choose randomly $h_{k}$ shall be compared in the experimental section \ref{sec:Experiments}.
   We shall assume some value $h_0 \in [h_{min} \, ; \, h_{max}] $ is provided, and we shall use the same convergence diagnosis as in the \shortname. 
 \begin{algo*}
  \centering
   \begin{subfigure}{0.5\linewidth}
\scriptsize
  \begin{algorithmic}
\While{Convergence is not attained}
\State Draw the index $i_k \in [n]$ uniformly

\State $\mathbf{x}_{i_k}^{(k+1)} \gets \mathcal{S}_h (\mathbf{x}_{i_k}^{(k)};\mathcal{X}^{(k)})$
    \State $k+1 \gets k$
    \State Convergence diagnosis
\EndWhile
\end{algorithmic}
\caption{Stochastic
  Mean-Shift}\label{alg:stochasticMeanShiftClustering}
  \end{subfigure}
   \begin{subfigure}{0.4\linewidth}
\scriptsize
  \begin{algorithmic}
\State choose any sequence $(\nu_k)$ s.t. $\nu_k \to 0$
\While{Convergence is not attained}
\State Draw the index $i_k \in [n]$ uniformly
\State Define $\delta=$  
$\min\lset \nu_k \, ; \, \lpar\frac{h_{k}}{h_{min}}\rpar^2 -1 \, ; \, 1-\lpar \frac{h_{k}}{h_{max}}\rpar^2 \rset $
\State Draw $\alpha \sim \mathcal{U}_{(1-\delta\, , \,1+\delta)}$
and define $h_{k+1} = \frac{h_{k}}{\sqrt{\alpha}}$
\State $\mathbf{x}_{i_k}^{(k+1)} \gets \mathcal{S}_{h_{k+1} } (\mathbf{x}_{i_k}^{(k)};\mathcal{X}^{(k)})$
    \State $k+1 \gets k$
    \State Convergence diagnosis
\EndWhile
\end{algorithmic}
\caption{Doubly Stochastic
  Mean-Shift}\label{alg:doubly_stochasticMSC}
  \end{subfigure}
  \caption{Stochastic and Doubly Stochastic mean-shifts.}
  \label{tab:DSMS}
\end{algo*}
 As can be seen: prior to the execution, we choose any non-negative sequence $(\nu_k )$ such that $\nu_k \to 0$ (though the  convergence rate is of no importance in the proofs, slow rates are preferred in practice); at each step $k$, given the BW $h_{k-1}$ from the previous step, we draw uniformly $\alpha \sim \mathcal{U}_{(1-\delta , 1+\delta )}$
 for $\delta = \min\lset \nu_k \, ; \, \lpar\frac{h_{k}}{h_{min}}\rpar^2 -1 \, ; \, 1-\lpar \frac{h_{k}}{h_{max}}\rpar^2 \rset $, which ensures two things: first, the new BW $\dsp h_{k+1} = \frac{h_{k}}{\sqrt{\alpha}}$ shall belong to 
 $[h_{min} \, ; \, h_{max}]$, and secondly $h_{k+1} - h_{k} \to 0$ as $k$ tends to infinity. In section \ref{sec:theory_with_proofs} we shall provide some reasons justifying such a way to define $(h_k)$. It should be precised that the sequences $(i_k ) , (h_k )$ are assumed to be independent.

 Stated equivalently: 
 after selecting randomly both an index $\ridx{k} \in [n]$ and a BW $h_{k+1}$ the $k+1^{\text{th}}$ state is defined as 
  \begin{align}\label{eq:step_dsms}
      \state^{(k+1)} = \state^{(k)} + \frac{h_{k+1}^{2} /2}{\sum\limits_{j=1}^n G\lpar \frac{\point_{\ridx{k}}^{(k)} - \point_{j}^{(k)}}{h_{k+1} }\rpar} \nabla_{\point_{\ridx{k}}} L_{h_{k+1}} (\state^{(k)}),
  \end{align}
 where $\nabla_{\point_{i}} L_{h_{k+1}} (\state)$ denotes the gradient of $L_{h_{k+1}}$ in the $\point_{i}$ direction.

 
\section{Theoretical results}
\label{sec:theory_with_proofs}

In this section we show that the \dsm provides almost surely a fixed clustering of the sample after a finite number of steps. 

Unless stated otherwise, in this section we suppose given any kernel $K$ defined on $\re^d $, associated to a profile $k$ satisfying all of the assumptions of Definition~1;  all notations, definitions, and assumptions introduced there will be used in the present section. The presented results' proofs are postponed to the \prooflocation.

\subsection{Submartingale property of \algName}
\label{subsec:theory_1}

Starting from an initial state $\state^{(0)} \in \re^{d n}$, i.e. an ordered set of $n$ points $ \state^{(0)}= [\point_{1}^{\top} , \ldots , \point_{n}^{\top} ]^{\top} $ such that $\point_{i} \in \re^d $ for each $i$, 
we already introduced the typical \dsm $k^{\text{th}}$ step in 
\eqref{eq:step_dsms}. 
 It is immediate that the state $\state^{(k)}$ depends both on the random sequence of indices $[\ridx{k-1},\ldots,\ridx{0}]$ and the random sequence of BW's $\lbr h_{k},\ldots,h_{1}\rbr$, i.e. there exists a non-random function $F^{(k)}: \lpar [n] \times [h_{min} \, ; \, h_{max}]\rpar^{k} \to \re^{d n}$ such that $\state^{(k)} = F^{(k)} [(\ridx{k-1},h_{k}) ; \ldots; (\ridx{0} ,h_{1})]$. 
 


 Fixing for a moment $h>0$, and supposing we perform the $k^{\text{th}}$ update step in the SMS, the following proposition (proved in~\cite{lapidot_stochastic_2025}) implies in particular that $ L_h \lpar \state^{(k+1)} \rpar > L_h \lpar \state^{(k)} \rpar$ provided $ \state^{(k+1)} \neq \state^{(k)}$:
  \begin{prop*}\label{prop:non_decreasing_L}
      Suppose the SMS with (fixed) BW $h$ updates $\state^{(k)}$ to $\state^{(k+1)}$, it holds that 
      \begin{align}\label{eq:non_decreasing_L} 
      L_h \lpar \state^{(k+1)} \rpar - L_h \lpar \state^{(k)} \rpar \geq \frac{2 |\kernprof^{\prime}(0)|}{h^2} \, \| \point_{\ridx{k}}^{(k+1)} - \point_{\ridx{k}}^{(k)} \|^2  
      \end{align}
      with $\ridx{k} \in [n]$ denoting the randomly chosen index at step $k$. We have also the following upper bound on the 
      gradient component along the direction $\point_{\ridx{k}}$:
       \begin{align}\label{eq:bounded_partial_gradient}
     \| \nabla_{\ridx{k}} L_h (\state^{(k)}) \| \leq \frac{2\, n  |\kernprof^{\prime}(0)|}{h^2 } \,\left \|  \point_{i_k }^{(k+1)} - \point_{i_k }^{(k)} \right \| \ ; 
 \end{align}
  \end{prop*}

  This ascending property of the $L_h $ values for a fixed $h$, combined to the particular way to draw randomly $h_{k+1}$ given the value $h_k$ which is exposed in algorithm \ref{alg:doubly_stochasticMSC}, has the interesting consequence that the sequence $\lpar L_{h_{k}} (\state^{(k)}) \rpar $ is a discrete-time (positive) submartingale as we shall see now. 

  \begin{prop*}\label{prop:submartingale_for_Lh}
      Let $0<h_{min}<h_{max}$ be two positive constants. Given $h_1 \in [h_{min} , h_{max}]$, suppose that for any positive integer $k$ we draw randomly $\dsp h_{k+1} = \frac{h_k }{\alpha^{\frac{1}{2}}}$ with $\alpha \sim \mathcal{U}_{(1-\delta , 1+\delta)} $ for $\delta=\delta_k $ defined as in algorithm \ref{alg:doubly_stochasticMSC}. Pick up randomly $i_k \sim \mathcal{U}_{[n]}$ and update the $i_{k}^{\text{th}}$ point $\point_{i_k }^{(k+1)} = \mathcal{S}_{h_{k+1} } (\point_{i_k}^{(k)};\state^{(k)})$ to define $\state^{(k+1)}$.       
      It holds a.s. that
      \begin{align} 
      \mathbb{E}\lbr L_{h_{k+1}}(\state^{(k+1)}) \ | \, \state^{(k)} \, ; \, h_k \rbr  \geq  L_{h_{k}}(\state^{(k)}) \label{eq:submartingale_for_Lh_values}
      \end{align}
      Thus the sequence $\lpar L_{h_{k}} (\state^{(k)}) \rpar $ is a discrete-time positive submartingale adapted to the process $\lpar \state^{(k)}\, ; \, h_k \rpar $.
  \end{prop*}

 \begin{rem*} As can be understood from the previous proof, given $h_k $ we could have drawn randomly $h_{k+1}$ in many other ways than in algorithm \ref{alg:doubly_stochasticMSC}, and still satisfy the same submartingale property. We used here this particular scheme for the sake of proof simplicity. \end{rem*}

  Since for any process sample $\omega $ one has $\forall \, k \in \nn \  L_{h_{k}(\omega)}(\state^{(k)}(\omega)) \leq \frac{n(n+1)}{2}\, 
  \kernprof(0)$, we may apply a Doob's convergence theorem~\cite{williams1991probability} to the positive submartingale $\lpar L_{h_{k}}(\state^{(k)}) \rpar $: in our case there exists a bounded non negative function $L_{\infty}(\omega)$ such that 
  $L_{h_{k}(\omega)}(\state^{(k)}(\omega)) \tend{k} L_{\infty}(\omega) $ almost surely on $\omega$. 

  At each step $k$, \eqref{eq:non_decreasing_L} and \eqref{eq:bounded_partial_gradient} above imply the following 
  \begin{align}
       \| \nabla_{\ridx{k}} L_{h_{k+1}} (\state^{(k)}) \|^2 & \leq 
       \frac{2 \, n^2 |\kernprof^{\prime}(0)|}{h_{k+1}^2 } \lbr L_{h_{k+1}} \lpar \state^{(k+1)} \rpar - L_{h_{k+1}} \lpar \state^{(k)} \rpar \rbr
        \label{eq:partial_gradient_bounding2} 
  \end{align}
 This has the following decisive consequence. 
 \begin{cor*}\label{cor:partial_gradient_tends_tozero_as}
  Keeping the same assumptions as in proposition \ref{prop:submartingale_for_Lh}, 
     given any initial state $\state^{(0)} \in \re^{d n}$:
     for any \dsm sample process it holds a.s. that $\nabla_{\ridx{k}} L_{h_{k+1}} (\state^{(k)}) \tend{k} \boldsymbol{0}$.  
 \end{cor*}


 This result will in turn imply the existence a.s. of a fixed clustering after a finite number of steps. 
 
  \subsection{Clustering and convergence results for \dsm}
\label{subsec:theory_2}
 Let us introduce convenient notations for the rest of this section. At step $k \in \mathbb{N}$ of the \dsm algorithm, the gradient $\nabla L_{h} (\state^{(k)})$ for a BW value $h>0$ shall be simply denoted by $\nabla^{(k)}_{h}$; hence $\nabla_{i\, ; \, h}^{(k)}$ shall denote $\nabla_{\point_i } L_{h } (\state^{(k)})$ for any $i \in [n]$; and $\| \nabla^{(k)}_{h } \|$ shall denote the norm $\max\limits_{i \in [n]} \| \nabla_{i\, ; \, h }^{(k)} \|_2 $ i.e the maximal component wise $\ell_2$ norm value of the gradient $\nabla^{(k)}_{h}$. 
 
 In this subsection we shall consider the process 
 $\lpar \state^{(k)} \, ; \, h_k \rpar $ which takes its values in the space $\re^{n d} \times [h_{min}\,;\,h_{max}]$. It is by our assumptions a non homogeneous Markov chain~\cite{meyn1993markov}. For any $k \in \nn $ we denote by $\alpha_k $ the random coefficient such that $h_{k+1} = h_k/\sqrt{\alpha_k }$, so that $h_{k+1}>h_{k} \iff \alpha_k < 1$. We introduce now definitions and notations that will be used in subsequent proofs.
  Suppose given any $\epsilon >0$ and fixed positive integer $k_0 $ ; we define the following event $\SmallSubg{k_0} = \lset \omega \, | \ \forall \, k \geq k_0 \ \| \nabla_{i_k \, ; \, h_{k+1} }^{(k)} (\omega) \|_2 < \epsilon \rset $. By previous considerations from subsection \ref{subsec:theory_1}, every process sample belongs a.s. to an $\SmallSubg{k_0}$ for some $k_0 \in \nn$. 
 
  For a given $\epsilon >0 $ as before, let us also introduce a sequence of stopping times $\lpar T_p \rpar_{p\in \nn}$ for the stochastic process $\lpar \state^{(k)} \, ; \, h_k \rpar $ which shall be used in the proofs. Let us define $T_1 = \min \lset k \in \nn \, | \, \| \nabla^{(k)}_{h_{k+1} } \| \geq \epsilon \rset $ (we set $T_1 = \infty$ is there is no such $k$);  for any $p \geq 2 $ we define $T_p = \min \lset k > T_{p-1} \, | \, \| \nabla^{(k)}_{h_{k+1} } \| \geq \epsilon \rset $ (set to $\infty$ if there is no such integer $k$). The sequence $\lpar T_p \rpar_{p\in \nn}$ is clearly adapted to the sequence $[(i_k \, ; \, h_{k+1})]_{k\in \nn}$ which  determines the process sample. 

  \begin{lem*} \label{lem:preliminary_to_prop_on_stopping_times}
      For any $\epsilon >0$, $k_0 \in \nn $ and $p \in \nn_0 $ we have 
      \begin{align}
          \mathbb{P}\lbr \lpar T_{k_0 + p} < \infty \rpar \SmallSubg{k_0} \rbr & \leq \lpar 1- \frac{1}{n}\rpar^p 
      \end{align}
  \end{lem*}
 
  The previous lemma has the following important consequence:
  \begin{prop*}\label{prop:nabla_Lh_goes_to_zero_as}
     Given any initial state $\state^{(0)} \in \re^{d n}$, and under the same assumptions as in proposition \ref{prop:submartingale_for_Lh}, the \dsm \,sequence satisfies 
      a.s. that $$\| \nabla^{(k)}_{h_{k+1} } \| \tend{k} \boldsymbol{0}_{d n}$$
  \end{prop*}

The main clustering result is now presented. It states that almost surely, after a specific step in the \shortname, well-separated and stable clusters are observed. 

    \begin{theorem*}\label{thm:clusters_existence}
    Under the same assumptions as in proposition \ref{prop:submartingale_for_Lh}, given any initial state $\state^{(0)} \in \re^{d n}$, let $\lpar \state^{(k)} \rpar $ be the \dsm process sample. Suppose $\tau$ is any given positive real s.t. $\tau<\frac{h_{min}}{2}$. 
        Then there exists a.s some partition $J_1 ,\ldots,J_M $ of $[n]$ 
        and $K \in \nn $ such that for every $k \geq K$:
        \begin{enumerate}[label=\arabic*$)$]
            \item $\forall \, l\in [M]$ it holds that 
            $\max\limits_{\substack{i \in J_l \\ j \in J_l }} \| \point^{(k)}_{i} - \point_{j}^{(k)} \| < \tau \ \ ; $ 
            \item $\forall\, (l,m) \in [M]^2 $ s.t $l \neq m$, it holds that
            $ \min\limits_{\substack{i \in J_l \\ j \in J_m }} \| \point^{(k)}_{i} - \point_{j}^{(k)} \| > h_{min}- \tau \ \ ; $
        \end{enumerate}
        in other words a.s. the points $\point_{1}^{(k)}, \ldots,\point_{n}^{(k)}$ belong to fixed clusters for sufficiently large $k$, 
        the $l^{th}$ cluster being defined as $\mathcal{C}_{l} = \lset \point_{i}^{(k)} \, , \, i \in J_l \rset $ (for all sufficiently large $k$); moreover all clusters diameters tend a.s to $0$ as $k \to \infty$.  
    \end{theorem*}
     
\subsection{Practical Convergence Diagnosis}
\label{subsec:practical_convergence_diagnosis}
 Given any initial state, the minimal number $k$ of \dsm steps which are necessary to achieve the aforementioned clustering is by nature a random variable. No theoretical results in this paper provide further precision about a higher bound on $k$ which would hold under a specified high probability. It has been pointed out that 
 each \shortname step costs linearly in $n$ since we shift only one point at a time~\cite{lapidot_stochastic_2025}, a property which is shared by \dsm; in addition, each \dsm step requires to draw randomly a bandwidth value. As a result, both algorithms, \ref{alg:doubly_stochasticMSC} and \ref{alg:stochasticMeanShiftClustering} might need more steps than
 their \bms or \mshift counterparts in order to achieve the desired
 clustering. This constraint makes it necessary to define convenient
 stopping criteria, even if the clustering above has not been
 completely achieved yet. In practice, whenever a sufficient number of points have a sufficiently small last shift, the algorithm stops. Such a practice is justified by the conclusions of propositions \ref{prop:non_decreasing_L} and \ref{prop:nabla_Lh_goes_to_zero_as}. 

\section{Numerical Experiments}
\label{sec:Experiments}

We present in this section the results obtained on synthetic data. The distribution of the samples and the clusters used were similar to \cite{lapidot_stochastic_2025}, for the sake of a fair comparison with the other algorithm. \dsm was compared with \mshift, \bms and \shortname. 

For quantitative examinations, We used the average number of clusters, the Average Cluster Purity (ACP) and the Average Label Purity (ALP), the latter being defined in \cite{Cohen2021,Ajmera2002,Manning2008} as:
\begin{align}
    ACP & = \frac{1}{Q} \sum\limits_{q=1}^{Q} \sum\limits_{r=1}^{R} P(d_r | c_{q})^2 \label{eq:ACP_def} \\ 
    ALP & = \frac{1}{R} \sum\limits_{r=1}^{R} \sum\limits_{q=1}^{Q} P(c_{q} | d_r )^2  \label{eq:ALP_def} 
\end{align}
where in the latter $P(d_r | c_{q})$ is the probability for a point of the cluster $c_q$ to have the true label $d_r$ and $P(c_{q} | d_r )$ the probability for a point with known label $d_r$ to belong to the cluster $c_q$, respectively. The inner sums in \eqref{eq:ACP_def} and \eqref{eq:ALP_def} defines a purity index for each individual cluster and label respectively.  Roughly speaking, a low $ACP$ indicates that the clusters are not class homogeneous, whereas low $ALP$ points to classes decomposed along several clusters. Taking both aspects into account can be done by considering the geometric average: 
 \begin{align}\label{def:K}
   K = \sqrt {ACP \cdot ALP} \ \ \in (0,1]
\end{align}

\subsection{Performance on Underrepresented Clusters}

In numerous applications, some clusters are not well represented. For example, in speaker diarization, the clustering task can be difficult for speakers who participate little to the recorded discussion~\cite{wang_speaker_2017}. In forensic speech processing, and specifically for short-duration speaker diarization, the system is required to distinguish between multiple speakers based on very few speech segments~\cite{imseng_robust_2009}. In these regimes, traditional algorithms are prone to overfitting noise and converging to spurious local modes, necessitating robust stochastic approaches that can infer stable clusters despite relying on little information~\cite{vinals_analysis_2019}.

In order to investigate the performance of \dsm in this framework, the following experiments were performed.
The data consists of a 2-dimensional Gaussian Mixture Model (GMM) with 3 isotropic clusters, whose parameters are summarized in Table~\ref{tab:gmm_params}.
\begin{table}[!ht]
    \centering
    \caption{Data generation hyperparameters for the 3-Gaussian Mixture Model. The values for Mean, Standard Deviation, and Sample Size are detailed for each Cluster ID.}
    \label{tab:gmm_params}
    \begin{tabular}{lccc}
        \toprule
        & \multicolumn{3}{c}{\textbf{Cluster ID}} \\
        \cmidrule(lr){2-4}
        \textbf{Parameters} & \textbf{Cluster 1} & \textbf{Cluster 2} & \textbf{Cluster 3} \\
        \midrule
        Mean Vector & $\begin{bmatrix} 1 \\ 1 \end{bmatrix}$ & $\begin{bmatrix} -1 \\ -1 \end{bmatrix}$ & $\begin{bmatrix} 1 \\ -1 \end{bmatrix}$ \\
        \addlinespace
        Covariance matrix & $0.65\mathbf{I}_2$ & $0.65\mathbf{I}_2$ & $0.65\mathbf{I}_2$ \\
        \addlinespace
        Points per Cluster ($N$) & $N \in [10;200]$ & $N \in [10;200]$ & $N \in [10;200]$ \\
        \bottomrule
    \end{tabular}
\end{table}
The number of points per cluster is variable to test the algorithm's performance on sparse data. 
Though this model might appear simplistic at first glance, it allows us to clearly identify the influence of the hyperparameters for each algorithm and to point out their most critical shortcomings.
The number of sample points was made equal for each cluster, and ranged from 10 to 200, in order to investigate the number of found clusters after convergence. The hyperparameters for each algorithm are summarized in Table~\ref{tab:algo_hyperparams}.
\begin{table}[!ht]
    \centering
    \caption{Hyperparameter settings for Mean-Shift (MS), Blurring Mean-Shift (BMS), Stochastic Mean-Shift (SMS), and Dynamic Stochastic Mean-Shift (DSMS).}
    \label{tab:algo_hyperparams}
    \begin{tabular}{lcccc}
        \toprule
        \textbf{Parameter} & \textbf{MS} & \textbf{BMS} & \textbf{SMS} & \textbf{DSMS} \\
        \midrule
        & \multicolumn{4}{c}{\textit{Common Parameters}} \\
        \cmidrule{2-5}
        Max Iterations & \multicolumn{4}{c}{$10^7$} \\
        Convergence Threshold & \multicolumn{4}{c}{$10^{-6}$} \\
        \cmidrule{2-5}
        & \multicolumn{4}{c}{\textit{Algorithm-Specific Parameters}} \\
        \cmidrule{2-5}
        Bandwidth  ($h$) & 0.6 & 0.6 & 0.6 & \textit{Variable} \\
        Bandwidth Range $[h_{min}, h_{max}]$ & N/A & N/A & N/A & $[0.2, 1.6]$ \\
        Limiting Bandwidth $h_{inf}$ & N/A & N/A & N/A & 0.6 \\
        $\nu_{k}$ Convergence Rate & N/A & N/A & N/A & $\dfrac{1}{\log_{10}(10+\log_{10} k)}$ \\
        \bottomrule
    \end{tabular}
\end{table}

\begin{figure}[!ht]
    \centering
    \includegraphics[width=0.95\linewidth]{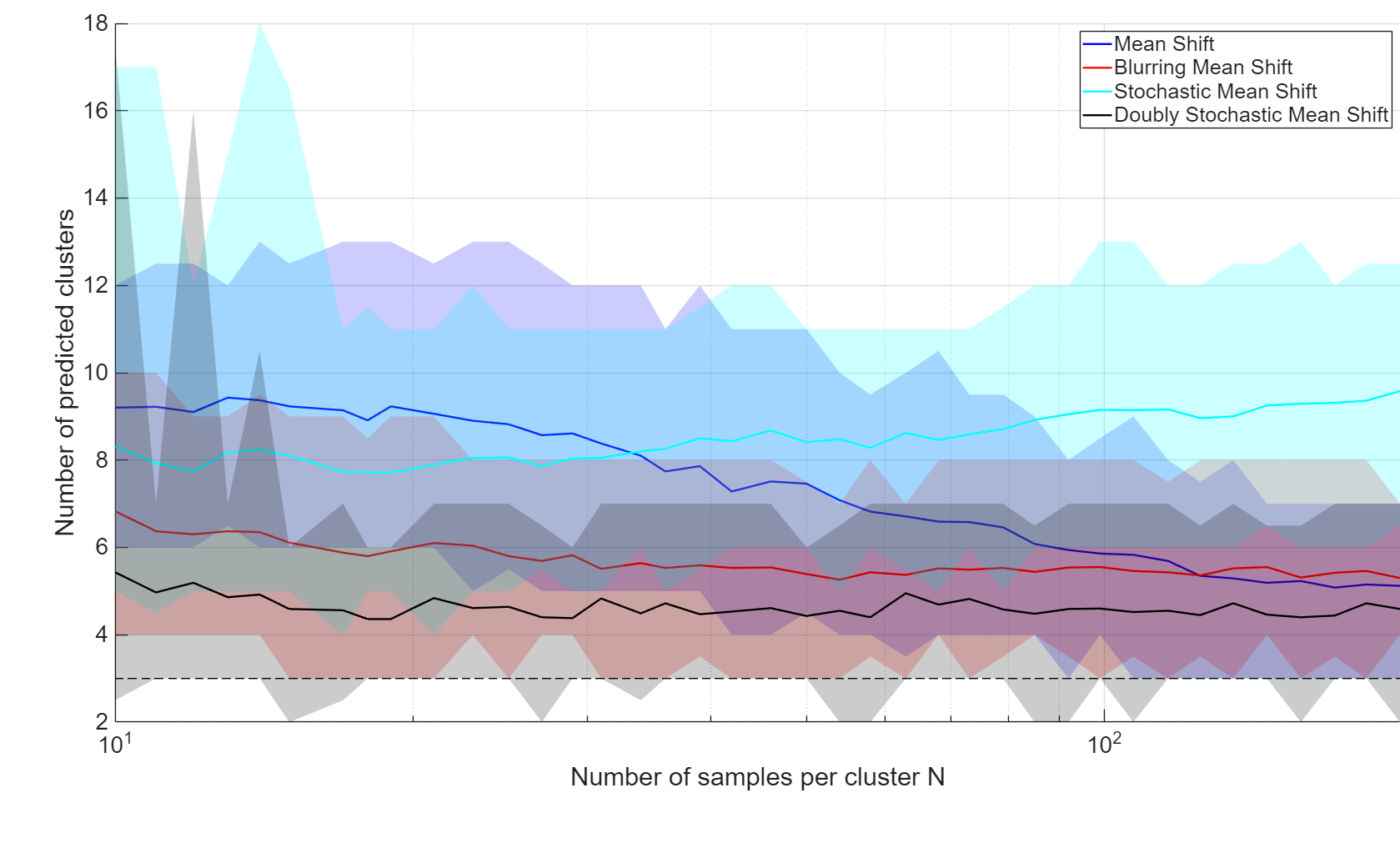}
    \caption{Average number of clusters with 90\% confidence intervals for the different algorithms, as a function of the number of samples per cluster. The true number of clusters (3) is represented by the thin dotted line.}
    \label{fig:sparse_clusters}
\end{figure}

For each setup, the experiments were performed 100 times. Figure~\ref{fig:sparse_clusters} illustrates the average number of clusters found using \mshift, \bms, \shortname, and \dsm, with their associated 90\% confidence interval.

We observe that DSMS largely outperforms SMS and MS, and provides a better cluster number estimate than BMS. These good performances for underrepresented clusters can be easily understood, as drawing a random bandwidth allows to regroup together sample points that other algorithms would consider outliers and leave unchanged. While the gap between the investigated methods is less significant when the sample size is sufficiently large ($N > 100$), a distinct difference emerges in the sparse regime ($10 \leq N \leq 50$). Indeed, \mshift and \bms exhibit a tendency toward \textit{over-segmentation} as the number of points per cluster decreases. This phenomenon is consistent with the limitations of fixed-bandwidth kernel density estimation, as scarce data can be interpreted falsely by a fixed bandwidth procedure as multiple modes, therefore increasing the number of resulting clusters artificially. \shortname, while more robust to local noise due to its asynchronous update rule, eventually suffers from the same limitation. The figure indicates that without an adaptive scale, stochasticity alone cannot bridge the significant density gaps present in small datasets. In contrast, by randomly varying both the selected point and the kernel size at each iteration, \dsm
 achieves improved performance by leveraging multi‑scale neighborhood information, maintaining a stable estimate of the cluster count across the entire range of sample sizes.
 Larger bandwidth samples allow the algorithm to traverse low-density regions that act as barriers for fixed-bandwidth methods, while smaller bandwidth samples refine the mode localization. An important point that will be highlighted by the following results is that this improvement does not degrade the clustering performance in terms of $K$.
 
  These observations also indicate the need to take the data repartition into account in the bandwidth sampling strategy, in order to optimize the clustering process. Data-dependent sampling strategies are out of the scope of the present paper and should be investigated in future contributions.

\subsection{Exhaustive Comparison with \shortname}
\label{subsec:comparison_with_sms}

\begin{figure}[!ht]
  \centering
  \includegraphics[width=0.8\linewidth]{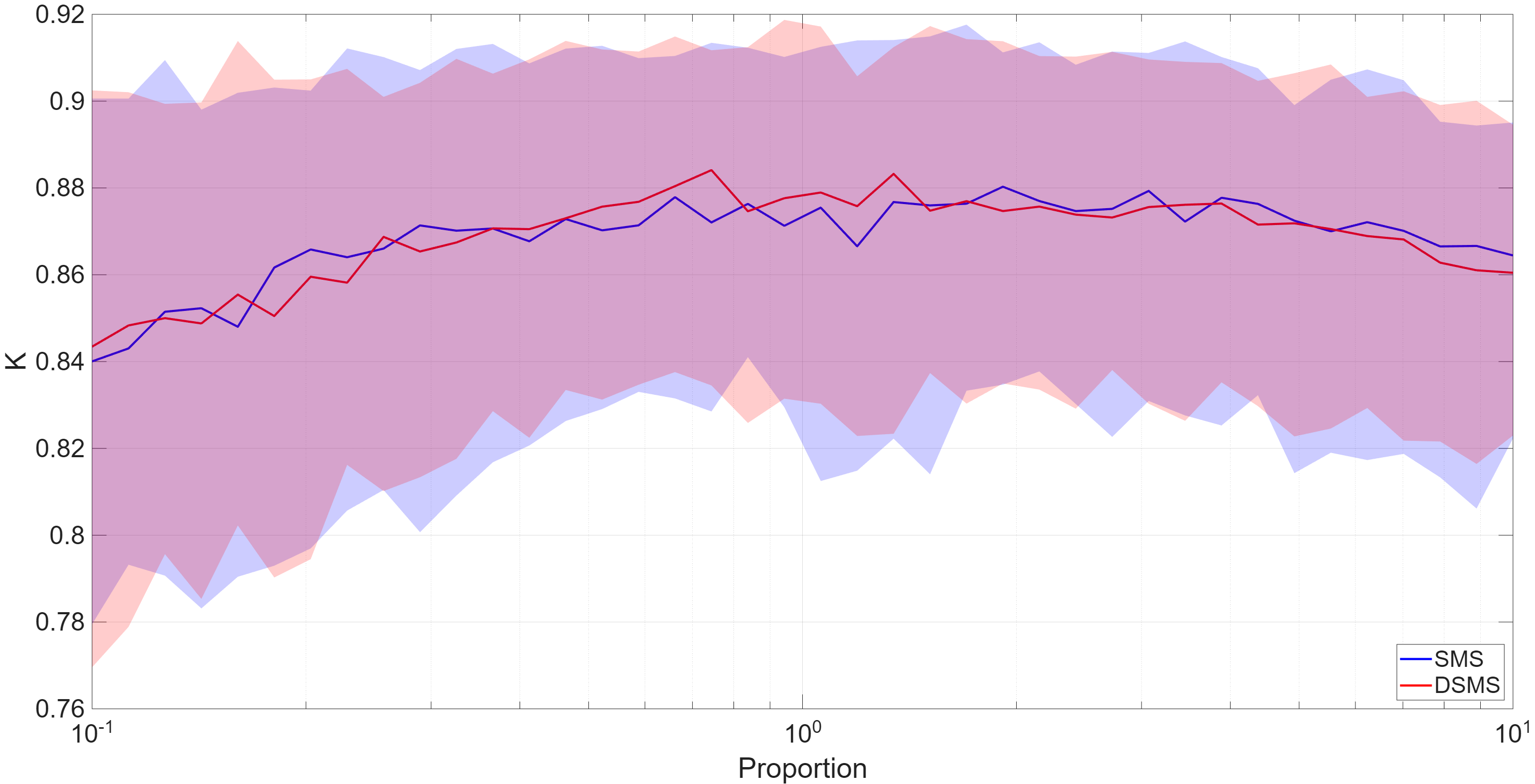}
  \caption{$K$ as a function of the class imbalance ratio as defined in~\cite{lapidot_stochastic_2025}}
  \label{fig:K_imbalance}
\end{figure}

\begin{figure}[!ht]
  \centering
  \includegraphics[width=0.8\linewidth]{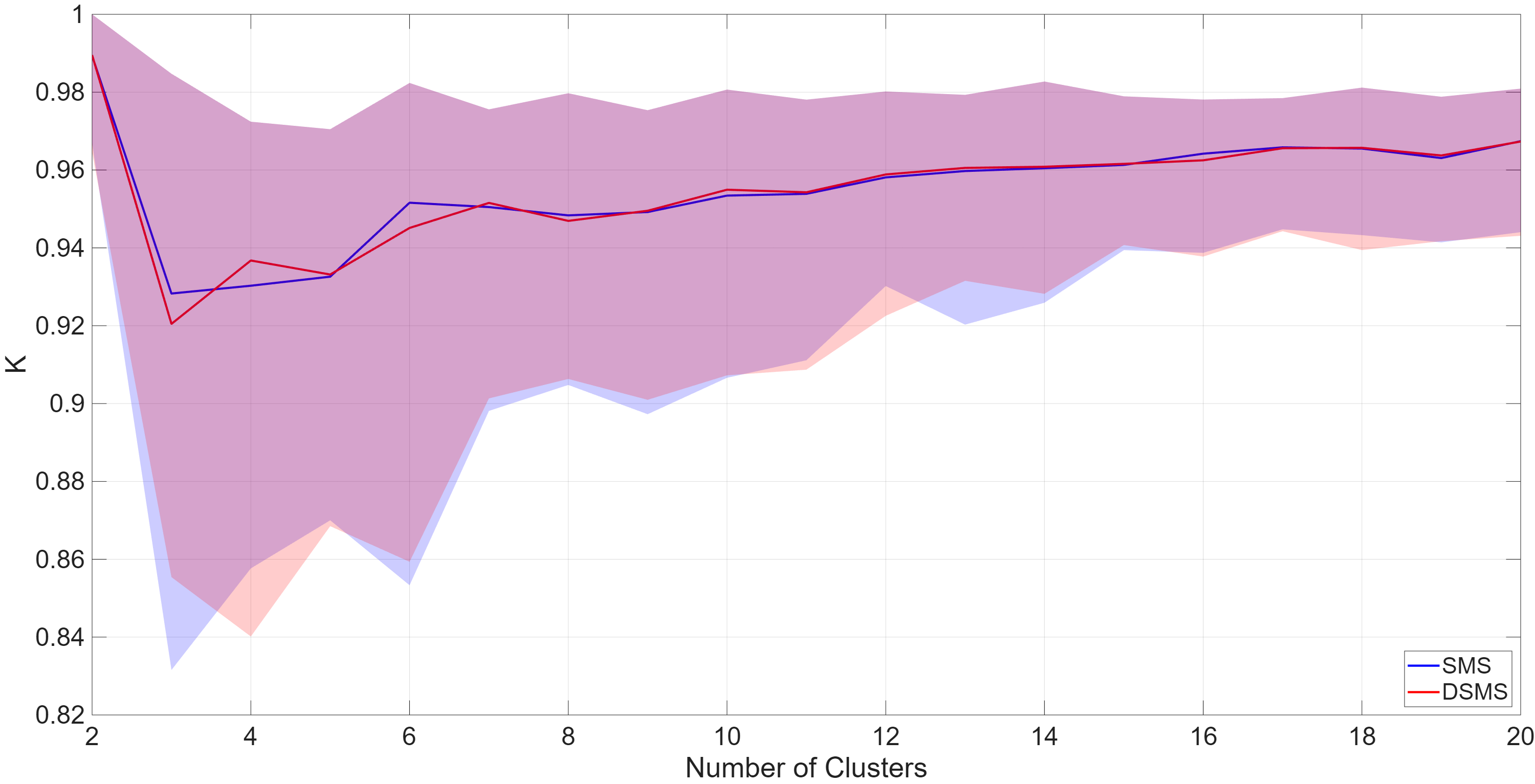}
  \caption{$K$ as a function of the number of clusters}
  \label{fig:K_nbclusters}
\end{figure}

The second experiment performed aimed to investigate whether a loss of performance is noticed when moving from \shortname to \dsm. To assess this question, we performed the same experiments as described in~\cite{lapidot_stochastic_2025} (we refer to the same publication for an in-depth description of the experimental setup and parameters' choice). These experiments evaluate performance of both \shortname and \dsm as a function of the number of clusters and the class imbalance. 
We present in Figures~\ref{fig:K_imbalance} and~\ref{fig:K_nbclusters}  
the evolution of $K$ for both \shortname and \dsm as a function of the ratio 
between clusters cardinals and, the number of clusters, respectively. All results are presented with their $90\%$-confidence interval.
%
%
The obtained results seem to demonstrate that provide the bandwidth range $h_{max} - h_{min}$ is conveniently chosen (see subsection \ref{subsec:radius_range} below), no loss of performance can be observed. Therefore, in such a case, \dsm should be preferred over its \shortname counterpart for better clustering results, as detailed previously.

\subsection{Influence of the bandwidth range}
\label{subsec:radius_range}
In algorithm \ref{alg:doubly_stochasticMSC} we gave one possible way to draw randomly the bandwidth sequences, which is both effective in the experiments and convenient for proving that a stable clustering is obtained almost surely after a finite number of steps. The global range to which the bandwidths belong is of great interest: indeed the slow $\nu_k $ convergence rate to $0$ provided in table \ref{tab:algo_hyperparams} assures that as $k$ tends to infinity, the bandwidths sequence $(h_k )$ will almost cover a whole range of values between two bounds $h_{min}$ and $h_{max}$ fixed by the user. In particular, and consistently with what we mentioned in subsection \ref{subsec:comparison_with_sms}, a key factor distinguishing \shortname from \dsm in terms of performance lies in the choice of the range $[h_{min} \,; \, h_{max}]$, and particularly of the width $h_{max}-h_{min}$. 

  Such a choice is a fundamental trade-off matter. First and briefly speaking, $h_{min}$ should not be too low and $h_{max}$ should be reasonably high, as drawing larger $h_k$'s facilitates the merging of fragmented components but risks over-smoothing (merging distinct clusters), whereas in contrast drawing low bandwidth values tends to restrict the algorithm's ability to bridge low-density regions, potentially leading to over-segmentation. Secondly, too low a bandwidth range $h_{max}-h_{min}$ hardly makes any difference with the \shortname. 

\begin{figure}
    \centering
    \includegraphics[width=0.8\linewidth]{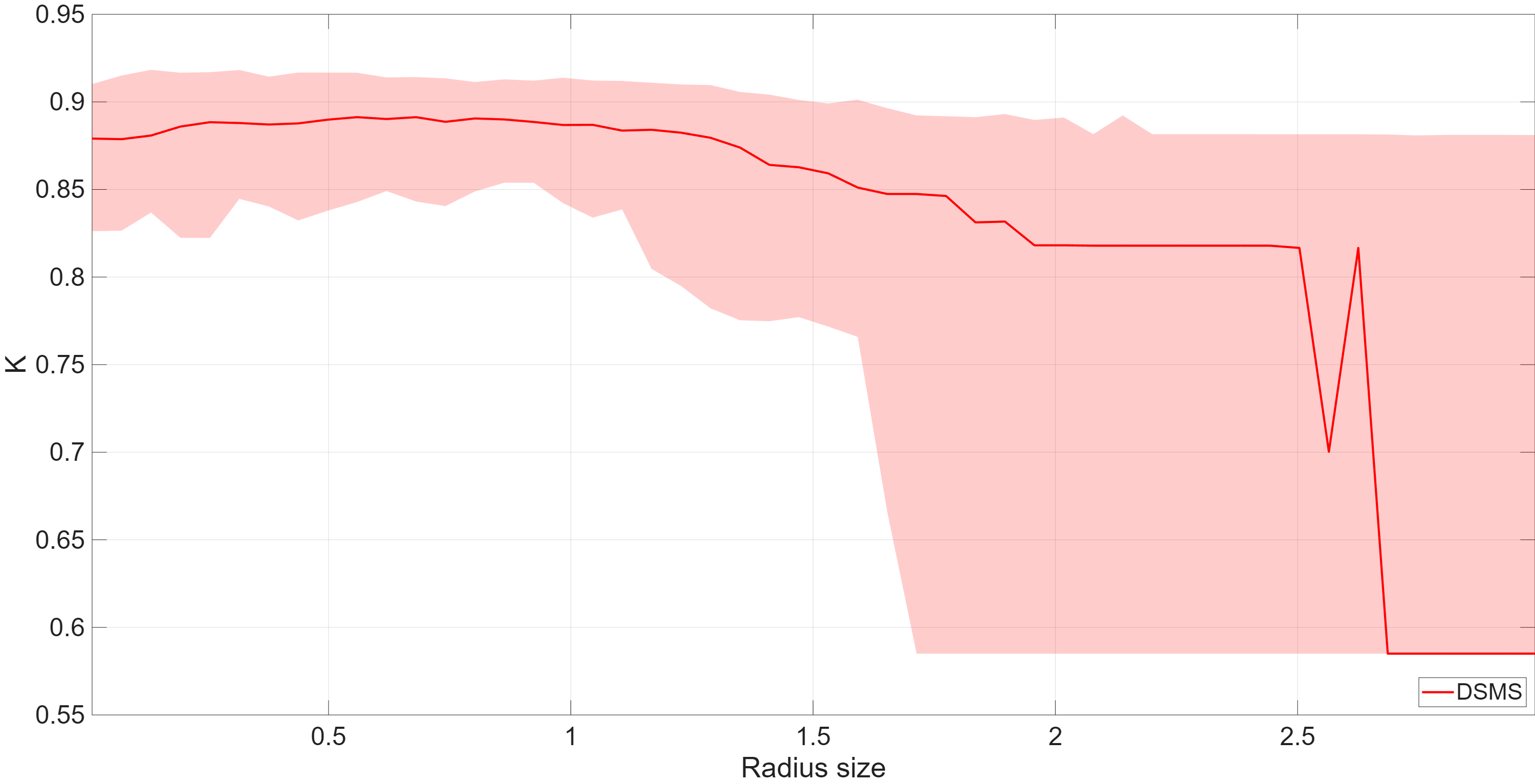}
    \caption{$K$ as a function of the bandwidth range $h_{max}-h_{min}$}
    \label{fig:K_radius}
\end{figure}

\begin{figure*}
    \centering
    \begin{subfigure}{0.48\textwidth}
        \includegraphics[width=\textwidth]{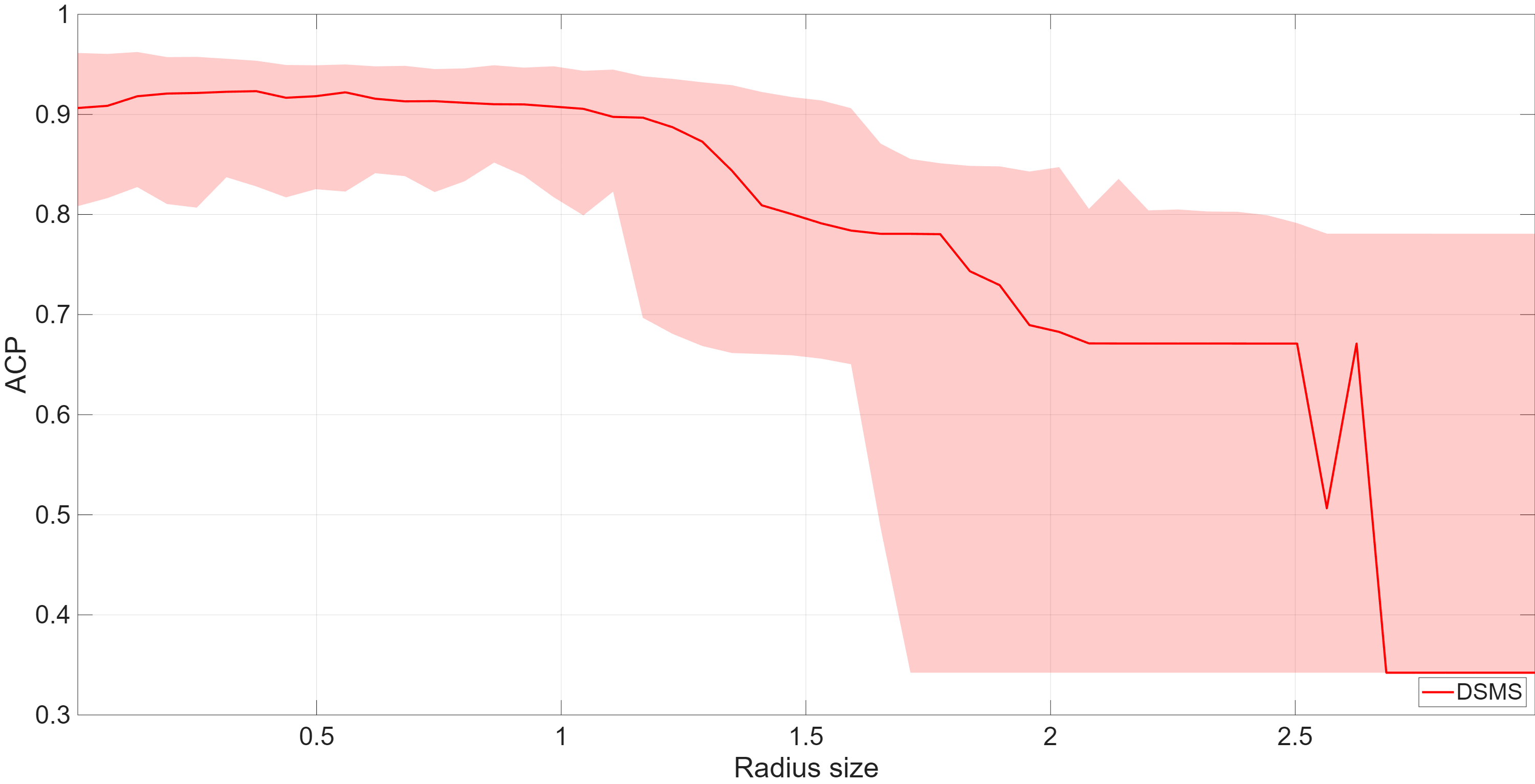}
        \caption{ACP}
        \label{fig:subim1}
    \end{subfigure}
    \hfill 
    \begin{subfigure}{0.48\textwidth}
        \includegraphics[width=\textwidth]{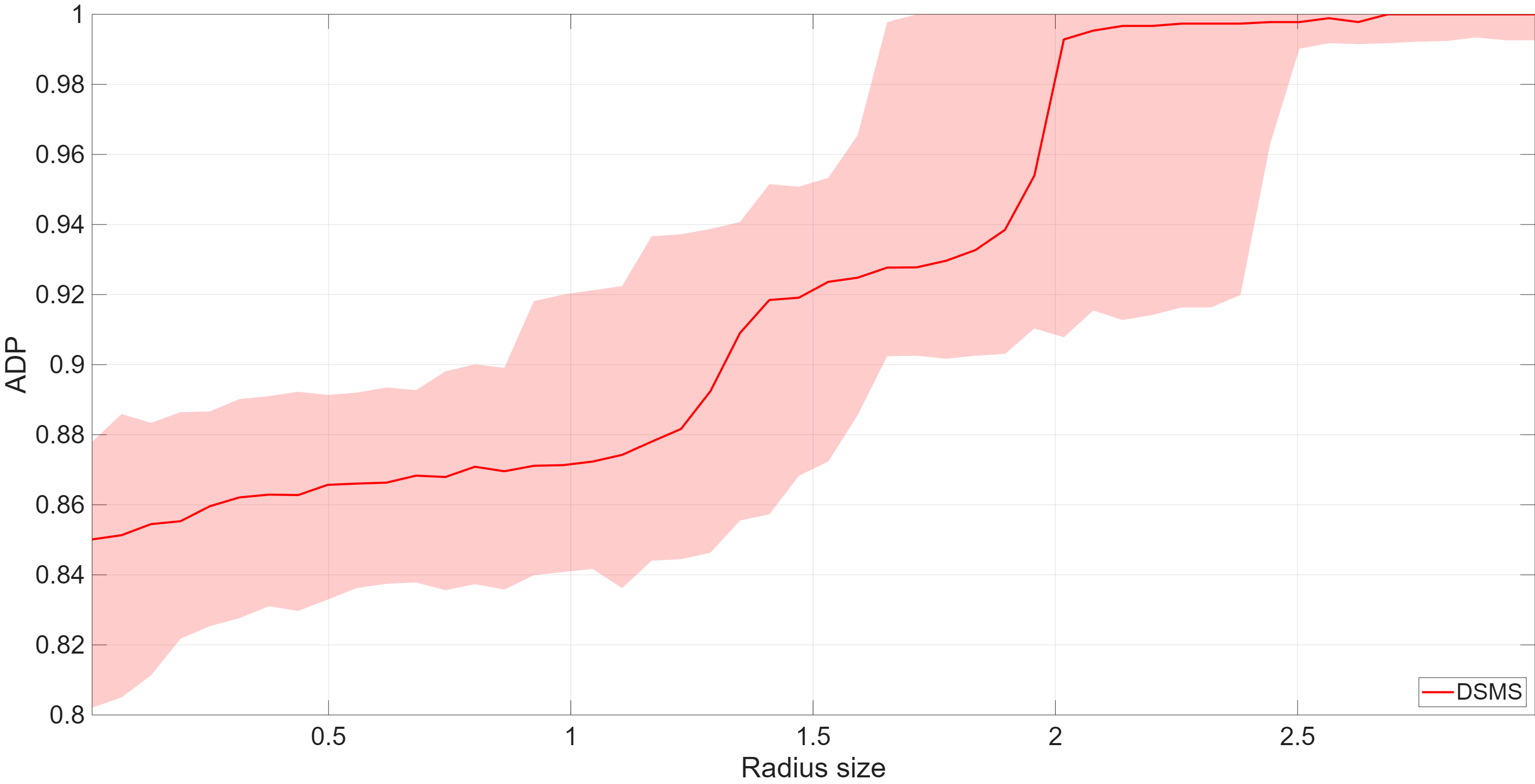}
        \caption{ADP}
        \label{fig:subim2}
    \end{subfigure}
    \caption{ACP and ALP as a function of $h_{max}-h_{min}$}
    \label{fig:ACP_ALP_radius}
\end{figure*}

We especially evaluated the influence of the bandwidth value on 100 runs of \dsm on 3 clusters with distribution similar as in Table~\ref{tab:gmm_params}. Each clusters included 100 data samples, and each run kept the data identical for all the changes in bandwidth range to avoid random artifacts. For each experiment, the limit bandwidth $h_\infty \in [h_{min},h_{max}]$ was chosen to get optimal results for \shortname.  Figure~\ref{fig:K_radius} presents the evolution of the parameter $K$ as the bandwidth range increases.
Not surprisingly, we observe a significant drop in performance as the bandwidth range increases. Notably, experiments indicate the existence of a range of radii so that better clustering performance than \shortname can be attained. This behavior can be further explained by the analysis of the ACP and ALP, presented in Figure~\ref{fig:ACP_ALP_radius}.

Indeed, as $h_{max}-h_{min}$ increases the algorithm's exploration capability improves, allowing it to pull outliers toward emerging clusters, and more generally to progressively merge groups which are likely parts of a same class. Consequently, the ADP (which serves as a proxy for completeness or label recall) increases. However, excessive merging inevitably corrupts the clusters with points from neighboring classes, leading to a degradation in ACP.

The crossing point of these opposing trends explains the stability region observed in the evolution of $K$. These results suggest that for specific topologies, introducing a reasonable amount of stochasticity in the choice of bandwidths yields a better balance between class separation on the one hand, and robustness to intra‑class variance on the other.
 It should be noted that the existence of some "optimal" range is intrinsically linked to the geometry and density of the data; characterizing this data-dependency remains an open challenge for future work.

\section{Conclusion}
\label{sec:Conclusion}

We presented \dsm, an improvement of the Stochastic Mean Shift algorithm where the radii around data samples are drawn randomly. Theoretical proofs including the existence a.s. of a stable clustering after a finite number of steps were presented, and results on synthetic dataset showed that this novelty enables a more accurate identification of sparse clusters, 
  without any degradation in its performance on the remaining criteria relative to \shortname.

 This work also indicates that intrinsic structure is rarely confined to a single scale, and that by integrating a random bandwidth policy, a known clustering process can be made more resilient to data scarcity and outliers. Further works will emphasize these aspects, and develop suitable sampling strategies towards optimality.

\appendix

\section{Proof of Proposition \ref{prop:non_decreasing_L}}

The proof of the first inequality can be found in see~\cite{lapidot_stochastic_2025}. The second assertion is an immediate consequence of the fact that $ \nabla_{i_{k} } L (\state^{(k)}) = \frac{2}{h^2} \sum\limits_{j=1}^n G\lpar \frac{\point_{i_k }^{(k)} - \point_{j}^{(k)}}{h}\rpar \lbr \state^{(k+1)} - \state^{(k)} \rbr $, taking into account that 
$G(\boldsymbol{u})= |\kernprof^{\prime}(\| \boldsymbol{u}\|^2 )| $ is a non increasing function of the radius $\| \boldsymbol{u}\|$. 

\section{Proof of Proposition \ref{prop:submartingale_for_Lh}}

  We start with the following inequality implied by proposition 
  \ref{prop:non_decreasing_L}:
  \begin{align}
      L_{h_{k+1}} (\state^{(k+1)}) - L_{h_{k}} (\state^{(k}) & = 
      L_{h_{k+1}} (\state^{(k+1)}) - L_{h_{k+1}} (\state^{(k}) \, + \, 
      L_{h_{k+1}} (\state^{(k)}) - L_{h_{k}} (\state^{(k}) \nonumber \\
      & \geq  L_{h_{k+1}} (\state^{(k)}) - L_{h_{k}} (\state^{(k}) \nonumber 
  \end{align}
 We shall now prove that 
 $\mathbb{E}\lbr L_{h_{k+1}}(\state^{(k)}) \ | \, \state^{(k)} \, ; \, h_k \rbr  \geq  L_{h_{k}}(\state^{(k)}) $. Let us denote for simplicity 
 $d_{i,j} = \| \point_{i}^{(k)} - \point_{j}^{(k)} \|$ for all $(i,j) \in [n]^2 $, then by definition 
 $\dsp L_{h_{k+1} }(\state^{(k)}) = \sum\limits_{1\leq i\leq j \leq n} \kernprof \lpar \frac{d_{i,j}^2 }{h_{k+1}^2 }\rpar $, hence using the definition of $h_{k+1}$ and Jensen inequality~\cite{rudin1987real} by the convexity assumption on the profile function $\kernprof$ we obtain the following:
 \begin{align}\label{eq:Jensen}
     \mathbb{E}\lbr L_{h_{k+1}}(\state^{(k)}) \ | \, \state^{(k)} \, ; \, h_k \rbr & = \sum\limits_{1\leq i\leq j \leq n} \, \int\limits_{(1-\delta\,;\, 1+\delta)}  \kernprof \lpar \alpha \frac{d_{i,j}^2 }{h_{k}^2 }\rpar \,\frac{1}{2\delta} d \alpha  \\
     & \geq \sum\limits_{1\leq i\leq j \leq n} \, \kernprof \lpar \frac{d_{i,j}^2 }{h_{k}^2 }\rpar = L_{h_{k}}(\state^{(k)})
 \end{align}
 since $\frac{1}{2\delta}\int\limits_{1-\delta}^{1+\delta} \alpha \, d \alpha = 1$. The proposition is proved.

\section{Proof of Corollary \ref{cor:partial_gradient_tends_tozero_as}}

Indeed \eqref{eq:partial_gradient_bounding2} above implies that for any $k \in \nn $:
 \begin{align}
      \| \nabla_{\ridx{k}} L_{h_{k+1}} (\state^{(k)}) \| 
        \leq \frac{n \sqrt{ 2 \,|\kernprof^{\prime}(0)|}}{h_{min} } \lbr L_{h_{k+1}} \lpar \state^{(k+1)} \rpar - L_{h_{k}} \lpar \state^{(k)} \rpar + \lpar L_{h_{k}} - L_{h_{k+1}} \rpar (\state^{(k)}) \rbr^{1/2} \label{eq:partial_gradient_bounding3}
  \end{align}
First it holds a.s. that $L_{h_{k+1}} \lpar \state^{(k+1)} \rpar - L_{h_{k}} \lpar \state^{(k)} \rpar \tend{k} 0$ by submartingale convergence. Moreover we do have $h_{k+1}-h_{k} \to 0$ by the way $(h_k)$ is obtained in algorithm 
\ref{alg:doubly_stochasticMSC}. At any step $k$ the updated point belongs to the convex envelope $\text{Conv}\lpar \state^{(0)}\rpar $ in $\re^d $ of the points composing $\state^{(0)}$ (by \eqref{eq:step_dsms} and straightforward induction), so by uniform continuity of the map $(\state, h) \mapsto 
L_h (\state)$ on the compact domain $\text{Conv}\lpar \state^{(0)}\rpar^n \times [h_{min},h_{max}]$ it follows that $\lpar L_{h_{k}} - L_{h_{k+1}} \rpar (\state^{(k)}) \to 0$ does hold. The corollary is proved. 

\section{Proof of Lemma \ref{lem:preliminary_to_prop_on_stopping_times}}

For any $k \in \nn $ we define the event 
 $B_k = \lset \| \nabla_{i_k \, ; \, h_{k+1} }^{(k)} (\omega) \|_2 < \epsilon \rset$. It clearly holds that 
 \begin{align}
 \mathbb{P}\lbr \lpar T_{k_0 + p} < \infty \rpar \SmallSubg{k_0} \rbr \leq 
 \mathbb{P}\lbr \lpar T_{k_0 < \infty} \rpar \cdots \lpar T_{k_{0}+p < \infty} \rpar B_{T_{k_0} } \cdots B_{T_{k_0 + p}} \rbr \label{eq:upper_bound_1}
 \end{align}
 Since any process sample $\omega$ is completely determined up to step $k$ by the sequence $\lbr (\ridx{k-1},h_{k}) ; \ldots; (\ridx{0} ,h_{1})\rbr$, for any positive integer $j$ we shall define $C_{j}$ as the event of all sequences $\underline{s} = \lbr h_{j+1} ; (\ridx{j-1},h_{j}) ; \ldots; (\ridx{0} ,h_{1})\rbr $ belonging to the events $T_{k_0 +p} = j$ and $ B_{T_{k_0} } \cdots B_{T_{k_0 + p-1}}$. We can write 
 \begin{align}
     \mathbb{P}\lbr \lpar T_{k_0 < \infty} \rpar \cdots \lpar T_{k_{0}+p < \infty} \rpar B_{T_{k_0} } \cdots B_{T_{k_0 + p}} \rbr & \leq 
     \sum\limits_{j=1}^{\infty} \int\limits_{\underline{s} \in C_{j} } \mathbb{P}(B_j \, |\, \underline{s}) \, d\text{P}(\underline{s})
 \end{align}
 Now for any $\underline{s} \in C_{j}$, the set $I_j (\underline{s}) = \lset i \in [n] \, | \,  \| \nabla_{i \, ; \, h_{j+1} }^{(j)} (\underline{s}) \|_2 < \epsilon\rset $ has cardinality $\leq n-1$ by assumption. By the assumption  $\ridx{j} \sim \mathcal{U}_{[n]}$ and the independence of $\ridx{j} $ w.r.t the sequence $\underline{s}$ it follows that 
 $\mathbb{P}(B_j  \, |\, \underline{s}) \leq 1-\frac{1}{n}$. Therefore we have
 \begin{align}
      \sum\limits_{j=1}^{\infty} \int\limits_{\underline{s} \in C_{j} } \mathbb{P}(B_j \, |\, \underline{s}) \, d\text{P}(\underline{s}) & \leq 
      \lpar 1- \frac{1}{n}\rpar \sum\limits_{j=1}^{\infty} \mathbb{P}(C_j ) \nonumber \\
      & = \lpar 1- \frac{1}{n}\rpar \mathbb{P}\lbr \lpar T_{k_0 < \infty} \rpar \cdots \lpar T_{k_{0}+p-1 < \infty} \rpar B_{T_{k_0} } \cdots B_{T_{k_0 + p-1}} \rbr \nonumber
 \end{align}
 By induction and \eqref{eq:upper_bound_1} it follows that 
 $  \mathbb{P}\lbr \lpar T_{k_0 + p} < \infty \rpar \SmallSubg{k_0} \rbr \leq \lpar 1- \frac{1}{n}\rpar^p $ as claimed. 

  \section{Proof of Proposition \ref{prop:nabla_Lh_goes_to_zero_as}}

   Given any $\epsilon >0$, lemma \ref{lem:preliminary_to_prop_on_stopping_times} implies immediately by letting $p \to \infty$ that for all $k_0 \in \nn $ we have $\mathbb{P}\lbr \lpar \forall \, p \in \nn \ T_p < \infty \rpar \, \SmallSubg{k_0}\rbr =0$.
  Since every process sample belongs a.s. to the chain union $\bigcup\limits_{k_0 \in \nn} \SmallSubg{k_0}$ by corollary \ref{cor:partial_gradient_tends_tozero_as}, it follows that 
  $$\mathbb{P}\lbr \forall \, p \in \nn \ T_p < \infty \rbr = \lim\limits_{k_0 \to \infty} \mathbb{P}\lbr \lpar \forall \, p \in \nn \ T_p < \infty \rpar \, \SmallSubg{k_0}\rbr =0$$
  In other words for each process sample there exists a.s. $K \in \nn $ such that $\forall \, k > K \ \| \nabla^{(k)}_{h_{k+1} } \| < \epsilon $. This is precisely the conclusion of the proposition.

\section{Proof of Theorem \ref{thm:clusters_existence}}
Let us denote as usual by $\omega = \lpar \state^{(k)} \, ; \, h_k \rpar $ a \dsm process sample. Consider the following event $\mathcal{B}$ (in the condition below $(k_p )$ denotes an infinite integers sequence): 
  \begin{align}
      \mathcal{B} = \lset \omega \,| \ \exists \, (k_p) \, \ \forall \, p \in \nn \, \, \exists (i,j) \in [n]^2 \ 
      \tau \leq \| \point_{i}^{(k_p )} - \point_{j}^{(k_p )} \| \leq h_{min}-\tau \rset \nonumber
  \end{align}
   Clearly $\mathcal{B} = \bigcup\limits_{(i,j) \in [n]^2 } \mathcal{B}_{i,j}$ with 
   \begin{align}
   \mathcal{B}_{i,j} = \lset \omega \,| \ \exists \, (k_p) \, \ \forall \, p \in \nn \, \ \tau \leq \| \point_{i}^{(k_p )} - \point_{j}^{(k_p )} \| \leq h_{min}-\tau \rset \nonumber 
   \end{align}
  Suppose $\omega \in \mathcal{B}_{i,j}$ for two given (and distinct) integers $(i,j)$, and let $(k_p)$ be an infinite sequence of integers satisfying the condition above; since the process takes its values in a compact set (as briefly justified in the proof of corollary \ref{cor:partial_gradient_tends_tozero_as}), one can further assume WLOG that $h_{1+k_p } \tend{p} h^{*}$ and $\state^{(k_p )} \tend{p} \state^{*}$ for 
  $h^{*} \in [h_{min},h_{max}]$ and $\state^{*} \in \re^{d n}$. 
  Since  $0<  \tau \leq \| \point_{i}^{*} - \point_{j}^{*} \| \leq h_{min}-\tau \leq h^{*} - \tau $ it follows from theorem \ref{thm:critical_point_charact} 
  that $\nabla_{h^{*}} L(\state^{*}) \neq \boldsymbol{0}_{d n}$, hence 
   $\| \nabla^{(k)}_{h_{k+1} } \| \not\longrightarrow \boldsymbol{0}_{n\,d}$ by the $C^1 $ assumption on the function $L_{h}(\state)$ w.r.t $(\state,h)$. 
  By proposition \ref{prop:nabla_Lh_goes_to_zero_as} it therefore follows that 
  $\mathcal{B}_{i,j}$ is of null measure, hence $\mathbb{P}\lpar \mathcal{B}\rpar = 0 $. 

  Therefore, it holds almost surely on $\omega $ that there exists
  $K= K(\omega) \in \nn $ such that the following two conditions hold forall 
  $k \geq K$:
  \begin{align}
    \forall \, (i,j) \in [n]^2 \ \, \| \point^{(k)}_{i} - \point_{j}^{(k)} \| & < \tau \ \, \vee \ \, \| \point^{(k)}_{i} - \point_{j}^{(k)} \| > h_{min} - \tau \label{thm_pf_cond1} \\
    \| \nabla^{(k)}_{h_{k+1}} \| & <  \frac{2 (h_{min}- 2 \tau)}{h_{max}^2}\, |\kernprof^{\prime}(0)| \label{thm_pf_cond2} \ \ \ ;
  \end{align}
  suppose now that two integers $(i,j)\in [n]^2 $ satisfies for some $k \geq K$ that $\| \point^{(k)}_{i} - \point_{j}^{(k)} \| < \tau $ and
  $\| \point^{(k+1)}_{i} - \point_{j}^{(k+1)} \| > h_{min}- \tau $ (or these two inequalities in opposite order, the argument will stay the same); since at any step the algorithm modifies one point only, we may assume that $\point_{j}^{(k+1)}=\point_{j}^{(k)}$ and therefore $ \| \point^{(k+1)}_{i} - \point^{(k)}_{i} \|>  h_{min}-2 \tau $ by the triangle inequality. 
  By \eqref{eq:step_dsms} on the other hand we must have 
  $\dsp \|  \point^{(k+1)}_{i} - \point^{(k)}_{i} \| \leq \frac{h_{max}^{2}}{2 |\kernprof^{\prime}(0)| } \| \nabla_{h_{k+1}}^{(k)} \| $, 
  which contradicts condition \eqref{thm_pf_cond2} above. 
  
   In other words 
  conditions \eqref{thm_pf_cond1}\eqref{thm_pf_cond2} above imply the existence 
  of some (fixed) partition $J_1 ,\ldots,J_M $ of $[n]$ satisfying a.s. conditions $1)2)$ in the theorem after some finite number of steps. Since $\tau$ was arbitrary positive the diameter of each cluster $\mathcal{C}_{l}, \, l \in [M]$ tends a.s to $0$ as $k \to \infty$.

\bibliographystyle{elsarticle-num-names}
\bibliography{mybib}








\end{document}